\documentclass[letterpaper, 10 pt, conference]{ieeeconf} 
\IEEEoverridecommandlockouts                             
\usepackage[utf8]{inputenc}
\usepackage[T1]{fontenc}
\usepackage{amssymb,amsmath,units,pgfplots,tikz}
\usepackage{accents} 			
\usepackage[hidelinks]{hyperref}
\usepackage[normalem]{ulem} 	
\usepackage[makeroom]{cancel} 	
\usepackage{makecell} 			
\usepackage{threeparttable}
\usepackage{mathtools} 			
\usepackage{enumerate}
\usepackage{varwidth}
\usepackage{algorithm,algpseudocode}
\floatname{algorithm}{Algorithm}

\newcommand{\maxn}[1]{_{\textrm{max},{#1}}}
\newcommand{\minn}[1]{_{\textrm{min},{#1}}}
\newcommand{\maxi}{_\textrm{max}}
\newcommand{\mini}{_\textrm{min}}
\DeclareMathOperator{\dt}{\mathrm{d}t}
\allowdisplaybreaks 
\title{\LARGE \bf Efficient Online Trajectory Planning for Integrator Chain Dynamics using Polynomial Elimination}
\author{Florentin Rauscher$^{1}$ and Oliver Sawodny$^{2}$
\thanks{$^{1,\,2}$ Institute for System Dynamics, University of Stuttgart, Germany (e-mail addresses: \{rauscher,sawodny\}@isys.uni-stuttgart.de)} }
\begin{document}
	\begin{titlepage}
		This work has been submitted to the IEEE for possible publication. Copyright may be transferred without notice, after which this version may no longer be accessible.
	\end{titlepage}
\maketitle
\thispagestyle{empty}
\pagestyle{empty}
\begin{abstract}
	Providing smooth reference trajectories can effectively increase performance and accuracy of tracking control applications while overshoot and unwanted vibrations are reduced. Trajectory planning computations can often be simplified significantly by transforming the system dynamics into decoupled integrator chains using methods such as feedback linearization, differential flatness or the controller canonical form. We present an efficient method to plan time optimal trajectories for integrator chains subject to derivative bound constraints. Therefore, an algebraic precomputation algorithm formulates the necessary conditions for time optimality in form of a set of polynomial systems, followed by a symbolic polynomial elimination using Gr\"obner bases. A fast online algorithm then plans the trajectories by calculating the roots of the decomposed polynomial systems. These roots describe the switching time instants of the input signal and the full trajectory simply follows by multiple integration. This method presents a systematic way to compute time optimal trajectories exactly via algebraic calculations without numerical approximation iterations. It is applied to various trajectory types with different continuity order, asymmetric derivative bounds and non-rest initial and final states.
\end{abstract}

\section{Introduction}
Control applications such as motion control of robotic systems, vehicles, construction machines or drive systems require the selection of appropriate reference signals during the transition between setpoints. Tracking performance and accuracy of feedback control can often be significantly improved by selecting the reference signals as a valid trajectory satisfying the essential plant dynamics. Furthermore, best results can usually be obtained by additionally providing the nominal input signal of a trajectory as feedforward control. Particularly the performance of linear feedback loops benefits from good reference signals since the input limitations of real-word systems typically require a conservative controller tuning. Many trajectory planning approaches complement feedback control properties well by allowing to take additional constraints into account such as input and state limitations or finite time convergence.
\begin{figure}
	\vspace{2mm}
	\includegraphics[width=.99\linewidth]{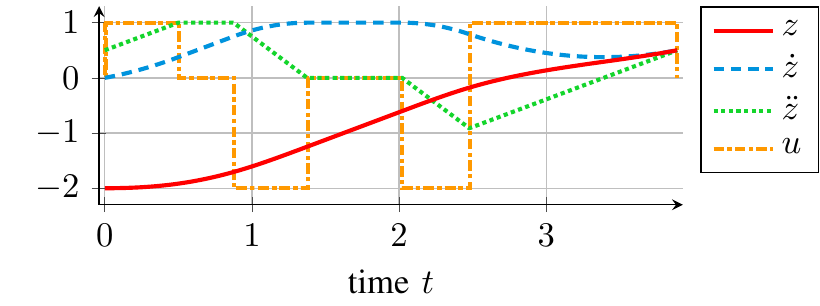}
	\vspace{-2mm}
	\caption{A time optimal trajectory for a chain of three integrators with nonzero initial and final conditions as well as derivative bound constraints.}
	\vspace{-5mm}
	\label{figc2trajectory}
\end{figure}

\subsection{Importance of integrator chain dynamics}
Trajectory planning for integrator chain dynamics plays an important role in practice. When the control tasks are not known beforehand, e.g. when reference values change regularly during set-point changes, trajectories must be planned or re-planned quickly once new reference values become available which requires fast algorithms. 
The primary goal is usually to generate smoothed $\mathcal{C}^k$ continuous reference signals with appropriately chosen continuity order~$k$ and bounded derivatives. This can be done efficiently by planning trajectories for chains of~$n=k+1$ integrators due to their simple dynamics and because derivative bounds for the reference can be represented as input and state bound constraints. Integrator chain dynamics also play an important role for plants that are controllable linear systems resp. feedback linearizable~\cite{Isidori2013} or differentially flat~\cite{Fliess1990} nonlinear systems. They can be transformed into controller canonical form resp. some analogue nonlinear normal form followed by an input transformation which allows to describe their dynamics by integrator chains. This simplifies trajectory planning since it leads to a decoupling into independent SISO channels and an exact linearization plus the simple differential relationship between the states. After planning, the trajectory is transformed back into the original coordinates so its states can be used as a reference for feedback and the input signals for feedforward control. For this trajectory generation strategy, the length of the integrator chains follows from the system order resp. the vectorial relative degree~\cite{Isidori2013}.

The simple structure of integrator chain dynamics can be exploited particularly well by algebraic planning methods. Their basic principle is to express the temporal progress of the output using elementary functions, so all states of the trajectory can be computed analytically as output derivatives. In the simplest case, equating their values at the start and final time with the desired initial and final states yields a system of equations that can be solved algebraically to determine the coefficients for the elementary functions. As a characteristic feature, algebraic methods typically lead to trajectories that converge in finite time. As another advantage all states and inputs are represented by elementary functions, so the computational effort does not depend on any time horizon length or time grid. Most algebraic planners can only consider input and state bounds of the integrator chain. However, physical plant constraints like torque limits can often be handled, e.g. using conservative integrator chain bounds such as a maximum acceleration and jerk or by scaling the resulting trajectory in time~\cite{Biagiotti2008}. Therefore algebraic trajectory planning plays an important role in practice.

\subsection{Related work}\label{secrelatedwork}
Many algebraic trajectory planning approaches use a polynomial ansatz for the output function. In the simplest case a transition time is specified as a design parameter so the polynomial of order $2^n-1$ with system order respectively integrator chain length~$n$ is computed based on the initial and final states via polynomial interpolation. A drawback of this simple approach is that the transition time must be specified and requires some heuristics or numeric search in order to satisfy constraints. Therefore, more sophisticated approaches have been developed. A comprehensive survey can be found in~\cite{Biagiotti2008}. Piecewise defined analytic functions, e.g. with trigonometric, exponential and higher order polynomial segments are useful when constraints such as maximum velocities must be taken into account. For example, piecewise polynomial functions with different order on each segment or combinations of polynomials, trigonometric functions and circular segments have been proposed in~\cite{Biagiotti2008,Bazaz1999,Macfarlane2003}.

Especially piecewise polynomials of minimal order play an important role. They consist of up to $2^n-1$ segments with an polynomial degree of~$n$ or lower. The input of the integrator chain changes between it's given bounds and zero (bang-zero structure), see Fig.~\ref{figc2trajectory}. For low system orders, such trajectories represent the time-optimal solution between the given initial and final conditions subject to bound constraints for the input and all integrator states~\cite{Ezair2014}. Various planning methods for trajectories of this type have been proposed like the numeric recursion as shown in \cite{Ezair2014,Knierim2012}. Starting with the $\mathcal{C}^{n-1}$ continuous signal of the first state, two~$\mathcal{C}^{n-2}$ continuous signals are planned for the second state such that the first state reaches a peak value that is found by a binary search or remains at its maximum value for some duration such that the first state reaches its target. The same is done recursively for the second and all further states which allows to efficiently compute trajectories also of higher order but not always yielding the time optimal solution. 
Also the generation of such trajectories via virtual variable structure control as a nonlinear filter has been proposed for third order trajectories~\cite{Zanasi2002,Biagiotti2010} which is very efficient for online generation when no full-horizon prediction is required.
 
Many planning methods for this trajectory type are based on algebraic equations and decision trees. They typically base on a computation of the switching time instants that can be used together with initial states and constraints to evaluate the trajectory in various ways. Many approaches focus on the third order case with general initial and steady-state final condition~\cite{Liu2002,Haschke2008,Besset2017} or set only the second derivative to zero at final time~\cite{Biagiotti2008,kroeger2010,Lambrechts2005,Jeong2005} which corresponds to the trajectory generator types III and IV when applying the naming scheme from~\cite{kroeger2010}. These planning methods have in common, that depending on which state constraints become active in a trajectory, a different underlying set of algebraic equations must be used for computation of the input switching times. As pointed out in~\cite{kroeger2010}, the main challenge is the development of rules or decision trees to select the corresponding set of equations such that for all admissible initial and final conditions a solution will be found. Since such decision trees can become very complex in the general case, most approaches are limited to low system orders and often only consider special cases with some initial or final state components being zero. Already the computations for the third order case with general initial and final states (type V in~\cite{kroeger2010}) are very complex and not covered by the aforementioned publications.

In order to overcome these limitations, the conditions for the trajectory can be formulated as polynomial systems and be solved algebraically using tools for polynomial elimination such as Gr\"obner bases~\cite{Buchberger1965}. They allow to transform the polynomial systems into triangular structure while preserving the same solution as the original one. This allows to solve them by sequential root search for univariate polynomials analogously to gaussian elimination for linear equation systems. The use of Gr\"obner bases for trajectory planning of integrator chains with input constraints has been proposed in~\cite{Walther2001} and with both input and state constraints in~\cite{Blaha2009,Blaha2014} for the case of three integrators. However, usability in practice is limited because the CPU-intensive polynomial elimination is done for each trajectory computation.

\subsection{New symbolic approach}
This paper presents a novel method for efficient online trajectory planning based on polynomial elimination that systematizes the algebraic approach. Therefore a symbolic precomputation is carried out that provides the general solution for any numeric values of constraints and initial or final conditions. It is based on a formulation of the trajectory conditions as multivariate polynomial systems in the switching times which are then decomposed into triangular structure. For efficient online trajectory planning, a computationally inexpensive algorithm is applied that evaluates the trajectory by finding the roots of the triangularized polynomial systems. This approach involves only algebraic computations, except for a numerical root search of high order polynomials, and can easily be adapted to simpler special cases with symmetries or without state constraints. Due to its computational efficiency it is suitable for practical use and has already been employed in several applications~\cite{Ringkowski2020GPobserver,Bauer2019,Rauscher2018,Rauscher2020}. 

This paper is organized as follows. In Section II, the symbolic precomputation is presented, followed by the online planning strategy in Section III including a discussion of the results. Finally, a conclusion is drawn in Section IV.

\section{Precomputation}
The dynamics of an $n$-integrator chain read 
\begin{align}
\dot{{x}} &= 
A
{x} 
+ 
B
u, \quad x(0)=x_0,\qquad \text{with}
\label{eqtrajsystem}
\\
A &=
	\begin{bmatrix}
	0 & 1  & & \\ 
	& 0  &\ddots & 1\\
	&  &\ddots & 0
	\end{bmatrix}
,\qquad 
B = 
	\begin{bmatrix}
	0 \\ \vdots \\ 1
	\end{bmatrix}
\end{align}
and state vector~${x}=\begin{bmatrix}x_{1},\dots,x_n\end{bmatrix}^\top\in\mathbb{R}^n$, input~$u\in\mathbb{R}$, time~$t\in\mathbb{R}$ as well as initial condition~$x_0$. The output of the integrator chain is denoted as~$z=x_1$, so the state components describe the derivatives~$(\frac{\mathrm{d}}{\dt})^i \,z = x_{i+1}$ and the input $(\frac{\mathrm{d}}{\dt})^n \,z = u$. 

The trajectories are planned as piecewise polynomials consisting of~$N=2^n-1$ segments with piecewise constant input~$u$ subject to derivative bound constraints for~$z$. The focus is on efficient computation of fast trajectories, not necessarily on time-optimality. Nevertheless, for low system orders $n\leq 3$ and in some situations also for $n>3$ they represent the solution of the time optimal control problem
\begin{align}
\underset{x,\,u}{\text{min}} \qquad &  T \label{ocp}\\
\text{s.t.}         	\qquad	& \dot{{x}} = A \,{x} + {B} \,u, \quad
{x}(0) = {x}_0, 	 \nonumber\\
& x\minn{i}  \leq {x}_i \leq  x\maxn{i},\quad i=2,\dots, n  \nonumber\\
& u\mini    \,\,\leq {u} ~\leq  u\maxi \nonumber,\\
& {x}(T) = {x}_\mathrm{f}
\nonumber
\end{align}
as argued in~\cite{kroeger2010,Ezair2014}. In this formulation the derivative constraints are written as state bounds $x\minn{i},\,x\maxn{i}$ and input bounds~$u\mini,\,u\maxi$, the desired final condition as~$x_\mathrm{f}$.

\subsection{Polynomial conditions for the trajectories}
In order to formulate the trajectories as piecewise polynomials, the switching times $t_i$,~$i\in \left[1,N\right]$ of the transitions into the following segment are defined with overall transition time~$t_N = T$ and $t_0=0$. The piecewise constant input~$u$ of the trajectory exhibits the characteristic basis pattern defined by
\begin{align}
u(t) = \begin{cases}
u\maxi 	& \text{if } \sigma_{i} = 1 	\\
u\mini	& \text{if } \sigma_{i} = -1 	\\
0		& \text{else} 					\label{inputtrajectory}
\end{cases}	\quad , \quad t_{i-1} \leq t < t_i.
\end{align}
with switching sequence~$\sigma$ according to algorithm~\ref{algSchaltmuster}. 
\begin{algorithm}[t]
	\caption{Switching sequence and constraint assignment}
	\label{algSchaltmuster}
	\begin{algorithmic}[1]
		\State {\textbf{input:} system order~$n$, first input sign~$\sigma_0$}
		\State {\textbf{output:} switching sequence~$\sigma$, constraint assignment~$\mathcal{B}$, }
		\State {$\sigma=\sigma_0$}
		\State {$\mathcal{B}=n+1$}
		\For   {$i\in\left[1,n-1\right]$} 
		\State {${\sigma} = \begin{bmatrix}{\sigma} & 0 & -{\sigma}\end{bmatrix}$}
		\vspace{1mm}
		\State {$\mathcal{B} = \begin{bmatrix}\mathcal{B} & (n-i+1) & -\mathcal{B}\end{bmatrix}$}
		\EndFor
		\State {$\mathcal{B} = \sigma_0\,\mathcal{B}$}
	\end{algorithmic}
\end{algorithm}
Whenever a state constraint is active, all higher derivatives including the input are zero. When no state constraints become active, the switching sequence consists of only~$n$ phases at maximum with~$n-1$ switchings and without any zero phases. This also holds for the time optimal control of all linear controllable systems and follows from the well-known Feldbaum theorem~\cite{feld1965optimal}. 
Every zero-phase of the input furthermore corresponds to an active constraint in a specific state component that can be identified by constructing the vector~$\mathcal{B}$ as shown in algorithm~\ref{algSchaltmuster}. The sign of the components in~$\mathcal{B}$ furthermore indicate if an upper or lower bound is active. The first sign~$\sigma_0$ of the sequence has to be determined individually for each trajectory. For example, the sequences for~$n=3$ read
\begin{align}
\sigma &= \begin{bmatrix}1&0&-1&0&-1&0&1\end{bmatrix},
\\
\mathcal{B} &= \begin{bmatrix}4& 3&-4& 2&-4&-3& 4\end{bmatrix}.
\end{align}
Input constraints are indicated by $\pm(n+1)=\pm~4$ in~$\mathcal{B}$. Overall, there are~$1+2+4\dots=\sum_{k=0}^n 2^k=2^n-1$ segments. Every second one, so overall~$n_\mathrm{x}=2^{n-1}-1$ segments, describes a state constraint. Since not all constraints become active in general, the segments can also degenerate to duration zero. For computation, the~$2^{n_\mathrm{x}}$ possible combinations of state constraints must be handled separately. In the following, they are denoted by a binary sequence in chronological order which contains a one for every corresponding state constraint that becomes active and a zero if not. 
\begin{figure}
	\centering
	\includegraphics[width=.9\linewidth]{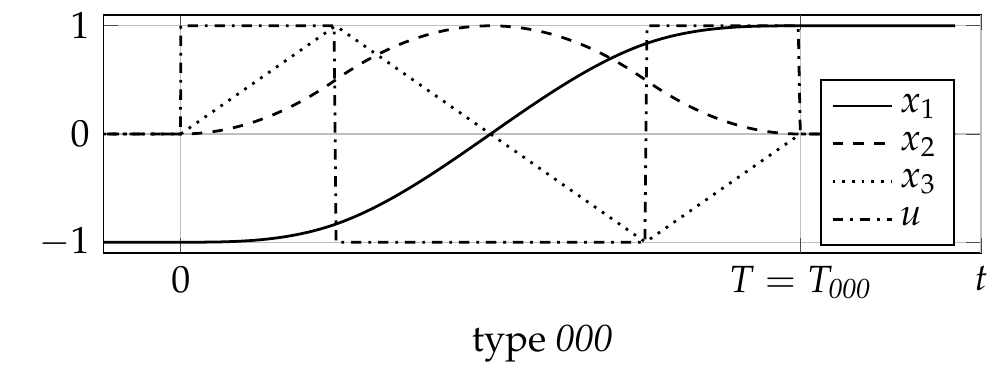}
	\includegraphics[width=.9\linewidth]{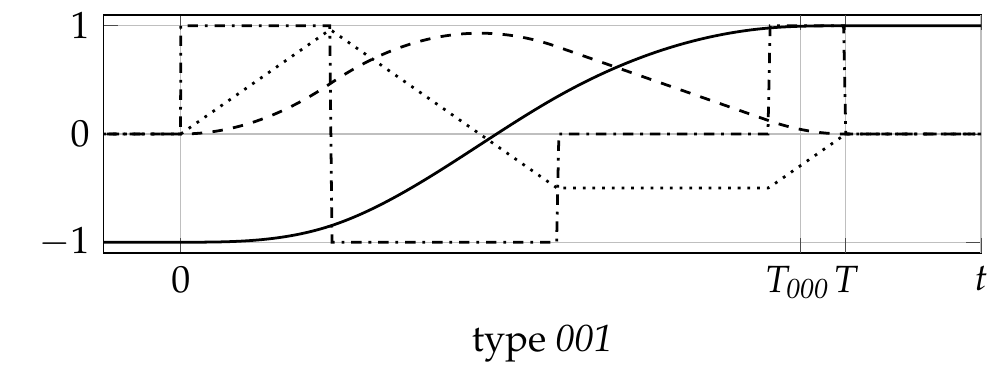}
	\includegraphics[width=.9\linewidth]{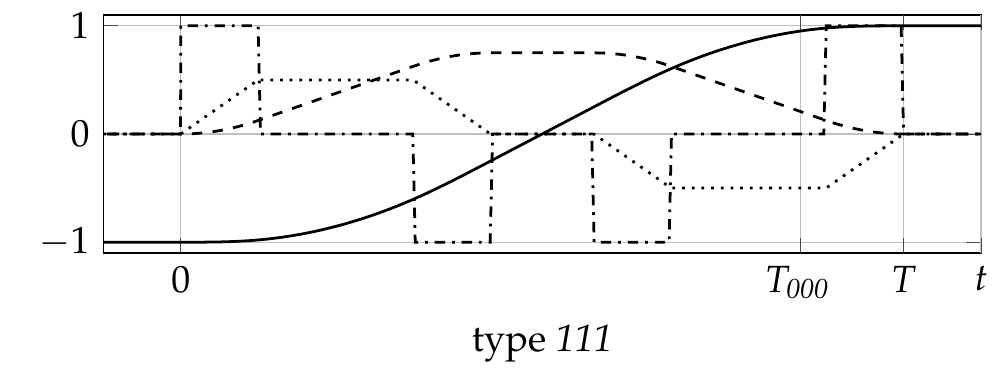}
	\vspace{-4mm}
	\caption{Exemplary trajectories of order~$n=3$ with different profile types of active state constraints.}
	\label{figpatterntypes}
	\vspace{-3mm}
\end{figure}
Fig.~\ref{figpatterntypes} illustrates three out of~$2^{n_\mathrm{x}}=8$ exemplary trajectories with different profile types for~$n=3$. The case~\textit{000} corresponds to the purely input-constrained case with shortest transition time~$T_\mathrm{\textit{000}}$, while for type~\textit{111} the state bounds are chosen tight enough to activate all possible constraints. For inactive constrains, the corresponding segment, and in some cases also the subsequent one, vanish (e.g. fifth segment of type \textit{001} in Fig.~\ref{figpatterntypes}). Particularly when repeatedly re-planning the trajectory, more segments vanish the closer it approaches the final state. 

The trajectory planning problem is based on three fundamental conditions:
\begin{enumerate}[I.]
	\item terminal condition
	\item state constraint conditions
	\item contact conditions.
\end{enumerate}
The components of the terminal condition
\begin{flalign}
\quad\textrm{I}.\quad 	x(T)=x_\mathrm{f}&& \label{eqterminalcondition}
\end{flalign}
ensure that the target state is reached. For each of the at maximum~$n_\mathrm{x}=2^{n-1}-1$ active state constraints, the trajectory reaches
\begin{flalign}
\quad\textrm{II}.\quad	x_i(t_j)=x\maxn{i} \quad  \text{resp.} \quad  x_i(t_j)=x\minn{i},\quad i\in\left[2,n\right]&& \label{eqconstraintcondition}
\end{flalign}
at the corresponding time which defines the state constraint conditions. Each state can reach an extremum at up to~$2^{k-2}$ segments. Active constraints for a state component, except for the last one, require that the higher derivatives become
\begin{flalign}
\quad\textrm{III}.\quad	x_{k}(t_j)=0,\qquad k\in\left[i+1,n\right] && \label{eqcontactcondition}
\end{flalign}
at this moment to ensure contact with the bound and no crossing. This introduces~$n-k$ contact conditions for each of the up to~$2^{k-2}$ state constraints, so~$\sum_{k=2}^{n-1}~2^{k-2}(n-k)$ in total. Overall, conditions I.-III. sum up to
\begin{align}
	n + (2^{n-1}-1) + \sum_{k=2}^{n-1}~2^{k-2}(n-k) = 2^n-1
\end{align}
conditions which can be shown by induction and equals the number of unknown switching times~$t_i$. Since for each inactive constraint one or more segments vanish but also the same number of switching times equal the previous one, the given conditions always equal the number of unknowns and allow to determine the trajectory algebraically.

Since it is not known \textit{a priori} which of the constraints become active, the possible profile types have to be iterated during computation. In worst case, the solution of all profile types has to be computed and the fastest one to be identified. By evaluating the types in the right order, i.e. beginning with the unconstrained case, the computation can be terminated preliminarily at certain points, if all constraints are satisfied. Note that the unconstrained trajectory represents the simplest case with fastest computation in this algebraic planning approach, while the trajectory with no inactive constraints represents the simplest case in the recursive approach of~\cite{Knierim2012}.

The conditions above are formulated as polynomial systems in the following with the switching times~$t_i$ as unknowns. For each profile type, one polynomial system is formulated whose number of polynomials equals the number of unknowns and depends on the combination of active constraints in the corresponding pattern. First, the polynomials are determined by multiple and segment-wise integration
\begin{align}
x = 
\begin{bmatrix}
\int_{0}^{t}\, \dots \int_{0}^{t} \, u \, \dt\dots\dt\\
\dots 	\\
\int_{0}^t\,  u \, \dt
\end{bmatrix}
,\quad x(0)=x_0\label{eqtrajalspiecewisepolynomial}
\end{align}
with $u$ as defined in~\eqref{inputtrajectory}. Note that all computations have to by done symbolically in order to keep the switching times, constraints and initial state components general.

Since all state components of the trajectory~\eqref{eqtrajalspiecewisepolynomial} result as piecewise polynomials, all terminal~(I.), state constraint~(II.) and contact conditions~(III.) can be formulated as the polynomial system resp. the set of polynomials
\begin{align}
\begin{bmatrix}
p_1(t_1,t_2,\dots)\\
p_2(t_1,t_2,\dots)\\
\vdots
\end{bmatrix}
 =& \begin{bmatrix}
 0\\0\\\vdots
 \end{bmatrix}
\label{eqPolynomsystem}
\end{align}
with multivariate polynomials~$p_i$ whose common positive real roots represent the switching time candidates. When multiple sets of positive roots fulfill all constraints, the fastest combination is selected to compute the trajectory.

\subsection{Polynomial elimination}
Since~\eqref{eqPolynomsystem} represents nonlinear, symbolic, multivariate polynomial systems, their algebraic solution is not straightforward. For trajectory planning, each solution set of positive real roots must be considered, so Newton-based numeric search methods that only find one root per polynomial and depend on the initial condition are no option. 

A better way to find all solutions are elimination methods for polynomial systems that have matured significantly in recent decades. This includes approaches such as Gr\"obner base methods~\cite{Buchberger1965}, Wu's characteristic set method~\cite{Wu1989} or regular chains~\cite{kalkbrener1993generalized}. Also a number of algebraic computation tools have been developed that facilitate polynomial elimination for various applications. In order to find the common roots of the polynomial conditions~\eqref{eqPolynomsystem}, their reduced Gr\"obner bases are computed which generate the same ideal but exhibit a triangular structure comparable to systems of linear equations after gaussian elimination. They can be seen as canonical forms of polynomial systems. The ideal generated by~\eqref{eqPolynomsystem} is zero-dimensional. Due to the triangular structure of each Gr\"obner basis, its roots can be found by elimination using iterative root search for univariate polynomials. Since efficient algorithms exist to find all solutions of univariate polynomials, this approach is suitable for trajectory planning.

In general, the symbolic polynomial systems~\eqref{eqPolynomsystem} contain singularities in the parameters, see example in Section~\ref{secexemplpolyelim}. They occur when divisors of some terms become zero, which can particularly happen for the important case of steady-state initial/final conditions. In order to handle these cases, for each profile type the divisor polynomials of all terms are determined and an auxiliary reduced Gr\"obner base is computed for each divisor polynomial. Therefore, the corresponding divisor is set to zero and solved for one parameter which is then plugged into the whole polynomial system and thereby reduces its complexity. During online trajectory computation, the divisor terms of each profile type are checked and, if one of them is zero, the corresponding auxiliary Gr\"obner base is evaluated instead of the default one. 

In order to check for constraint violations, the values of the state polynomials are evaluated at the segments in which a constraint is possible according to~$\mathcal{B}$, but not active. If those values are all within the bounds and the initial and final conditions are feasible, the trajectory is valid. Even when the initial conditions are within the specified bounds they might not be feasible, i.e. there might not exist a valid bounded input signal to prevent the trajectory from crossing the boundary constraints. E.g., when starting with~$x_{0,2}=x\maxn{2}$ and ~$x_{0,3}=x\maxn{3}$. For such infeasible cases, a solution of the polynomial systems will still be found that shortly violates the corresponding constraint, but then reaches the final state without further constraint violations. This is useful in practice, where small numeric errors can render an initial condition infeasible, but a trajectory that slightly violates a constraint might still be better than no solution. The same holds for final conditions that can not be reached without constraint violations, e.g. $x_{\mathrm{f},2}=x\maxn{2}$ and~$x_{\mathrm{f},3}=x\minn{3}$.

The basic structure of the precomputation is outlined as Algorithm~\ref{algpre}. 
\begin{algorithm}[t]
	\caption{Precomputation}
	\label{algpre}
	\begin{algorithmic}[1]
		\State {\textbf{Numeric Input:} $n$}
		\State {\textbf{Symbolic Input:} $x_0,\,x_\mathrm{f},\,x\maxi,\,x\mini,\,u\maxi,\,u\mini$}
		\State {\textbf{Output:} function files of polynomial coefficients}
		\For   {For each profile type}
		\State {formulate~\eqref{eqterminalcondition}-\eqref{eqcontactcondition} as polynomial system~\eqref{eqPolynomsystem}}
		\State {define relevant switching times~$t_i$}
		\State {compute reduced Gr\"obner basis of \eqref{eqPolynomsystem}}
		\State {compute auxiliary Gr\"obner bases for singular cases}
		\State {determine evaluation order for elimination}
		\State {determine polynomial coefficients for elimination}
		\State {save polynomial coefficients as function files}
		\EndFor
	\end{algorithmic}
\end{algorithm}
After symbolic computation of the reduced Gr\"obner bases for each profile type the evaluation order and the corresponding unknowns are determined. In general, the evaluation order of the unknowns does not equal the chronological order of the switching times. This allows to save the coefficients of the univariate polynomial in the Gr\"obner basis as a function of the initial/final states and constraints as input. The second polynomial of the Gr\"obner basis is also considered as an univariate polynomial, but its coefficients additionally depend on the roots of the first polynomial and so on. Therefore, the reduced Gr\"obner bases are saved as function files describing the coefficients of univariate polynomials which can then be evaluated numerically online for trajectory planning.

As a major advantage of symbolically solving~\eqref{eqPolynomsystem} using Gr\"obner bases the computations for polynomial elimination only have to be made once for each system order and can efficiently be employed for planning trajectories with any valid initial and final conditions or constraints. Special cases such as symmetric constraints, some zero states in initial or final conditions or further prior knowledge about the trajectories to be planned should directly be included in the polynomial systems~\eqref{eqPolynomsystem} which can drastically simplify the Gr\"obner bases and speedup trajectory planning significantly.

For the general case and increasing system order, the polynomial expressions become quite large and the computations time- and memory-consuming. So for high order trajectories, other planning approaches might be more suitable. But at least for system order up to~$n=3$ the general case can be fully solved using elimination software available today and fast online planning can be realized. Input-constrained trajectories with general initial and final conditions can be planned at least for up to~$n=4$. Rest-to-rest trajectories between steady states with symmetric bounds are solvable algebraically up to~$n=6$. With this ability, algebraic trajectory planning is already applicable to many use cases in practice. The computations are limited by symbolic elimination software, CPU power and memory, so trajectories of higher order are presumably computable in future.

\subsection{Exemplary polynomial elimination}\label{secexemplpolyelim}
\begin{figure}
	\includegraphics[width=.99\linewidth]{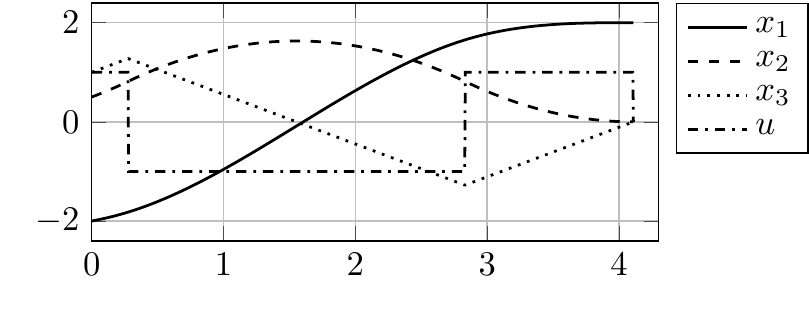}
	\vspace{-4mm}
	\caption{Trajectory of order~$n=3$, profile type \textit{000} (no state constraints active), with non-zero initial condition and symmetric input constraints from example in Section~\ref{secexemplpolyelim}.}
	\label{figC2example}
			\vspace{-5mm}
\end{figure}
In the following, the polynomial elimination for trajectories of order~$n=3$ is exemplarily shown for profile type \textit{000} (no state constraints active) with general initial conditions but steady state final conditions and symmetric constraints. First, the terminal condition is formulated based on the piecewise polynomial representation \eqref{eqtrajalspiecewisepolynomial} of the trajectory. The terminal state components read
\begin{align}
x_{\mathrm{f},1} =
&\,u\maxi\,\,\left(
\frac{t_1^3+t_7^3-t_6^3}{6}
+ \frac{t_1\, t_7^2+t_6^2\, t_7-t_1^2\, t_7-t_6\, t_7^2}{2} 
\right) \nonumber\\
& + u\mini \left(
\frac{t_2^2\, t_7+t_3\, t_7^2+t_4^2\, t_7+t_5\, t_7^2-t_2\, t_7^2}{2} 
\right) \nonumber\\
& + u\mini \left(
\frac{-t_3^2\, t_7-t_4\, t_7^2-t_5^2\, t_7}{2} 
+
\frac{t_5^3 + t_3^3 - t_2^3 - t_4^3}{6} 
\right) \nonumber\\
& + x_{0,1} + x_{0,2}\, t_7 + \frac{x_{0,3}\, t_7^2}{2} \, , 		\label{eqtrajpoly111x1}
\\
x_{\mathrm{f},2} =
&\,\,u\maxi\,\left(
- \frac{t_1^2}{2} + t_1\, t_7 + \frac{t_6^2}{2} - t_6\, t_7 + \frac{t_7^2}{2}
\right) 
\nonumber\\
&+ u\mini \, \left(- t_7\, t_4 - \frac{t_5^2}{2} + t_7\, t_5\right)
+ x_{0,2} + t_7\, x_{0,3} \nonumber\\
& + u\mini \, \left(
\frac{t_2^2}{2} - t_7\, t_2 - \frac{t_3^2}{2} + t_7\, t_3 + \frac{t_4^2}{2} 
\right) 
\\ 
x_{\mathrm{f},3} =&\,\,	u\maxi\, \left(t_1 - t_6 + t_7\right) +  u\mini\, \left(t_3 - t_2 - t_4 + t_5\right) + x_{0,3} .
\label{eqtrajpoly111x3}
\end{align}
For profile type~\textit{000}, the four segments corresponding to the state constraints vanish, since they are inactive. Therefore only the switching times $t_1$, $t_3$ and $t_7$ have to be determined as unknowns, while the other ones follow as
\begin{align}
t_1=t_2,\quad t_3=t_4=t_5=t_6. \label{eqtrajpoly000vereinf}
\end{align}
Conditions II. and III. do not apply for this profile type. For other profile types, the corresponding state constraint conditions (II.) have to be selected from 
$x_3(t_2) = x\maxn{3}$, 
$x_2(t_3) = x\maxn{2}$ and 
$x_3(t_5) = x\minn{3}$, eventually all with inverted sign. 
In the case of an active constraint for~$x_2$ also the contact condition (III.) $x_3(t_3)= 0 $ holds.

The reduced Gr\"obner basis for profile type \textit{000} can be computed using symbolic elimination software as
\begin{align}
0=&\,
2\, u\maxi\, t_1   + \frac{3 \, u\maxi\, \left(4\, u\maxi\, x_{0,2} - x_{0,3}^2\right)}{8\, u\maxi\, x_{0,2} - 4\, x_{0,3}^2}\, t_7 
\nonumber\\&
- \frac{96\, u\maxi^2\, (x_{\mathrm{f},1} - x_{0,1}) + 7\, {x_{0,3}}^3 + 12\, u\maxi\, x_{0,2}\, x_{0,3}}{24\, u\maxi\, x_{0,2} - 12\, x_{0,3}^2} 
\nonumber\\&
+
 \frac{u\maxi^3}{8\, u\maxi\, x_{0,2} - 4\, x_{0,3}^2}t_7^3 
+ \frac{3\, u\maxi^2\, x_{0,3}}{8\, u\maxi\, x_{0,2} - 4\, x_{0,3}^2} t_7^2
\!\label{eqtraj000stufenform1}
\\
0=&\,
t_3 
+ \frac{u\maxi^2\,t_7+3\, u\maxi\, x_{0,3}}{16\, u\maxi\, x_{0,2} - 8\, x_{0,3}^2}\,t_7^2
\nonumber\\&
+ \frac{96\, u\maxi^2\, (x_{0,1}- x_{\mathrm{f},1}) + 5\, {x_{0,3}}^3 - 36\, u\maxi\, x_{0,2}\, x_{0,3}}{24\, u\maxi\, \left(2\, u\maxi\, x_{0,2} - x_{0,3}^2\right)} 
	\nonumber\\&
+ \frac{\left(x_{0,3}^2 + 4\, u\maxi\, x_{0,2}\right)}{16\, u\maxi\, x_{0,2} - 8\, x_{0,3}^2}\, t_7
,\label{eqtraj000stufenform2}
\\
0=&\,
t_7^4 
+ \frac{4\, x_{0,3}}{u\maxi}\, t_7^3 
+ \frac{ \left(16\, u\maxi\, x_{0,2} - 2\, x_{0,3}^2\right)}{u\maxi^2}\,t_7^2
\nonumber\\&
- \frac{48\, u\maxi^2\, x_{0,2}^2 + {x_{0,3}}^4 + 96\, u\maxi^2\, x_{0,3}\, (x_{\mathrm{f},1}-x_{0,1})}{3\, u\maxi^4} 
	\nonumber\\&
- \frac{ \left(96\, u\maxi^2\, (x_{\mathrm{f},1} - x_{0,1}) + 4\, {x_{0,3}}^3\right)}{3\, u\maxi^3} \,t_7
.\label{eqtraj000stufenform3}
\end{align}
It can be seen that only the last polynomial~\eqref{eqtraj000stufenform3} is univariate. So the solutions of this fourth order polynomial in~$t_7$ have to be computed first and then the positive real solutions are iteratively plugged into \eqref{eqtraj000stufenform2} and~\eqref{eqtraj000stufenform1} to find the corresponding solutions for $t_3$ and~$t_1$. Among all sets of solutions satisfying $t_1 \leq t_3 \leq t_7$ the fastest one is returned. While the input constraint value can be assumed~$u\maxi\neq0$, the singularity when $x_{0,3}^2 = 2\, u\maxi\, x_{0,2}$ (e.g. steady state initial condition) must be handled by an auxiliary reduced Gr\"obner basis. This is done by substituting~$x_{0,2}=\frac{x_{0,3}^2}{2\,u\maxi}$ in~\eqref{eqtrajpoly111x1}-\eqref{eqtrajpoly111x3} for which the much simpler reduced Gr\"obner basis
\begin{align}
0=&\, 2\, u\maxi\, t_1 - \frac{u\maxi}{2}\,t_7 +	\frac{3\, x_{0,3}}{2} \label{eqgroebnersingulara}
\\ 
0=&\, 	t_3 - \frac{3}{4}\, t_7 + \frac{x_{0,3}}{4\, u\maxi}
\\ 
0=&\,	
\frac{3\, u\maxi^3\, t_7^3 + 9\, u\maxi^2\, x_{0,3}\, t_7^2  + 9\, u\maxi\, {x_{0,3}}^2\, t_7 }{3\, u\maxi^3} \\
&\quad	+ \frac{96\, u\maxi^2\, (x_{0,1} - x_{\mathrm{f},1}) - 13\, {x_{0,3}}^3 }{3\, u\maxi^3} 
\label{eqgroebnersingularc}
\end{align}
can be computed that does not contain further singularities. Also the polynomial degree of the univariate polynomial reduced from four to three. 

Consider the example of a third order trajectory from initial state~$x_0=\left[-2,\,\frac{1}{2},\,1\right]^\top$ to final state~$x_\mathrm{f}=\left[2,\,0,\,0\right]^\top$ subject to only the input bounds~$|u|\leq u\maxi=1$. Here, the solution is a type \textit{000} profile. Since the singularity is active, the parameters are plugged into~\eqref{eqgroebnersingulara}-\eqref{eqgroebnersingularc} which yields
\begin{align}
	0=&\, t_1 - \frac{1}{4}\,t_7 +	\frac{3}{4} 
	\\ 
	0=&\, 	t_3 - \frac{3}{4}\, t_7 + \frac{1}{4}
	\\ 
	0=&\,	t_7^3 + 3\,  t_7^2  + 3\,  t_7 + \frac{-397}{3}.
\end{align}
Only the last polynomial is univariate and therefore solved first. It's only positive real solution is $t_7\approx4.11~\mathrm{s}$ which is plugged into the other polynomials yielding $t_3\approx 2.83~\mathrm{s}$ and $t_1\approx 0.28~\mathrm{s}$. The trajectory is simply evaluated by plugging these switching times and~$x_0$ into~\eqref{eqtrajalspiecewisepolynomial} for time values~$0\leq t \leq t_7$. The result is shown in Fig.~\ref{figC2example}.

\section{Trajectory planning}
\subsection{Computing the switching times}
Based on the reduced Gr\"obner bases, trajectories can be planned efficiently. This is done by iteratively computing the solutions of the Gr\"obner bases for each profile type and selecting the fastest one that satisfies all given constraints according to Algorithm~\ref{algtraj}. 
\begin{algorithm}[t]
	\caption{Trajectory planning}
	\label{algtraj}
	\begin{algorithmic}[1]
		\State {\textbf{Inputs:} $x_0,\,x_\mathrm{f},\,x\maxi,\,x\mini,\,u\maxi,\,u\mini$}
		\State {\textbf{Outputs:} switching times $t_i,\,i\in [1,2^n-1]$, trajectory coefficient tensor~$P$}
		\For   {each profile type}
		\State {select files describing default Gr\"obner basis}
		\If	   {singularity}
		\State {switch to auxiliary Gr\"obner basis}
		\EndIf
		\State \begin{varwidth}[t]{\linewidth}
			determine all solutions $t_i$ of Gr\"obner basis by iter-\par
			atively computing all roots of univariate polynomials
		\end{varwidth}
		\vspace{-.0mm}
		\State {find fastest solution satisfying all constraints, if any}
		\EndFor
		\State {select fastest solution satisfying all constraints}
		\State {compute piecewise polynomial coefficents for all state components of trajectory}
		\State {collect these polynomial coefficents in tensor~$P$}
	\end{algorithmic}
\end{algorithm}
If the default Gr\"obner basis of a profile type has divisors that are zero, the corresponding auxiliary basis has to be used instead. In order to solve the triangular polynomial systems represented by a Gr\"obner basis for the unknown switching times, the roots are iteratively computed for each polynomial. By starting with the univariate one and iteratively plugging each of its roots into the other polynomials, another one becomes univariate and can be solved.

When iterating the profile types in suitable order beginning with only active constraints in higher derivatives, the for-loop can be terminated after certain numbers of for-loop iterations if a valid solution has already been found. E.g., when starting with the purely input-constrained profile type, further iterations can already be skipped if it generates a trajectory that stays within the state bounds. 

Once the best combination of switching times $t_i$ has been determined, the trajectory~\eqref{eqtrajalspiecewisepolynomial} as a piecewise polynomial over time can easily be evaluated based on the initial condition~$x_0$. It can be returned as a tensor~$P$ that describes the polynomial coefficients as a vector for each state component and for each segment. Once switching times and coefficient tensor have been computed, the trajectory $(x,\,u)$ can efficiently be evaluated for arbitrary time values~$t$.

For low system orders~$n$ the univariate polynomials of the reduced Gr\"obner bases are of low polynomial degree. E.g. for~$n=3$ in the general case and lexicographic monomial order, the resulting systems consist of polynomials with degree $\leq 4$ for which all roots can be calculated algebraically. Hence, for up to~$n=3$ our approach allows to plan trajectories using purely algebraic calculations. 
For higher system orders~$n$, not all polynomial roots can be found algebraically. Also general guaranteed upper degree bounds for Gr\"obner bases are usually much higher than the maximal polynomial degree of the original polynomial system~\cite{moller1984upper}. In order to find all roots of higher degree polynomials, they can be determined as eigenvalues of the companion matrix, which usually involves some numeric iterations.

It should be noted that theoretically numerical problems can appear in Algorithm~\ref{algtraj} since the polynomial coefficients in~$P$ depend on previously computed roots of other polynomials and might therefore exhibit small numerical errors, e.g. originating by loss of significance effects. As shown in~\cite{wilkinson1963}, finding the roots of polynomials with approximate coefficients can be numerically unstable. However, this effect does not pose a major problem in the context of this trajectory planning approach, presumably because of the low polynomial degrees appearing in the Gr\"obner bases. For safety reasons, every trajectory should still be checked for accuracy against given tolerances.

\begin{figure}
	\includegraphics[width=.99\linewidth]{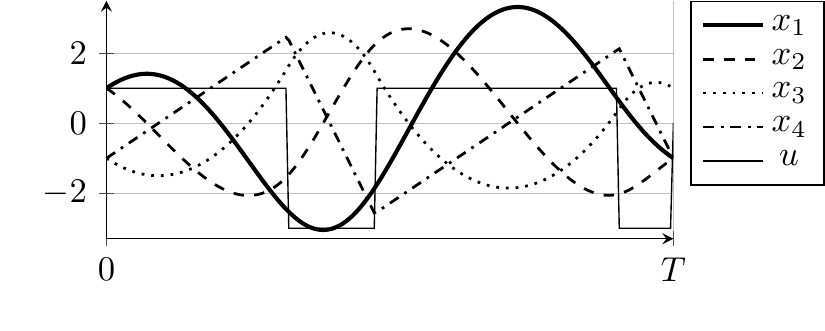}
	\vspace{-6mm}
	\caption{Fourth order trajectory with non-zero initial and final conditions and asymmetric input constraints.}
	\label{figC3u}
	\vspace{-3mm}
\end{figure}
\begin{figure}
	\includegraphics[width=.99\linewidth]{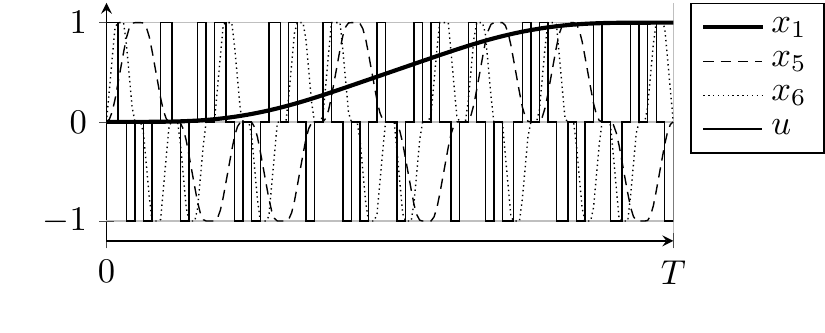}
	\vspace{-6mm}
	\caption{Sixth order trajectory with steady state initial and final conditions as well as symmetric state and input constraints (state variables normalized).}
	\label{figC5}
	\vspace{-5mm}
\end{figure}
\subsection{Results and implementation}
\subsubsection{Exemplary trajectories}
Some exemplary trajectory types are shown in the following for which the symbolic precomputation has been successful. The general approach is also valid for higher order trajectories but computational effort increases rapidly, as discussed above. The planning approach has been implemented in Matlab\textsuperscript{\textregistered} using symbolic math toolbox for the precomputation and automatic C code generation for efficient trajectory planning. 

\begin{figure}
	\includegraphics[width=.99\linewidth]{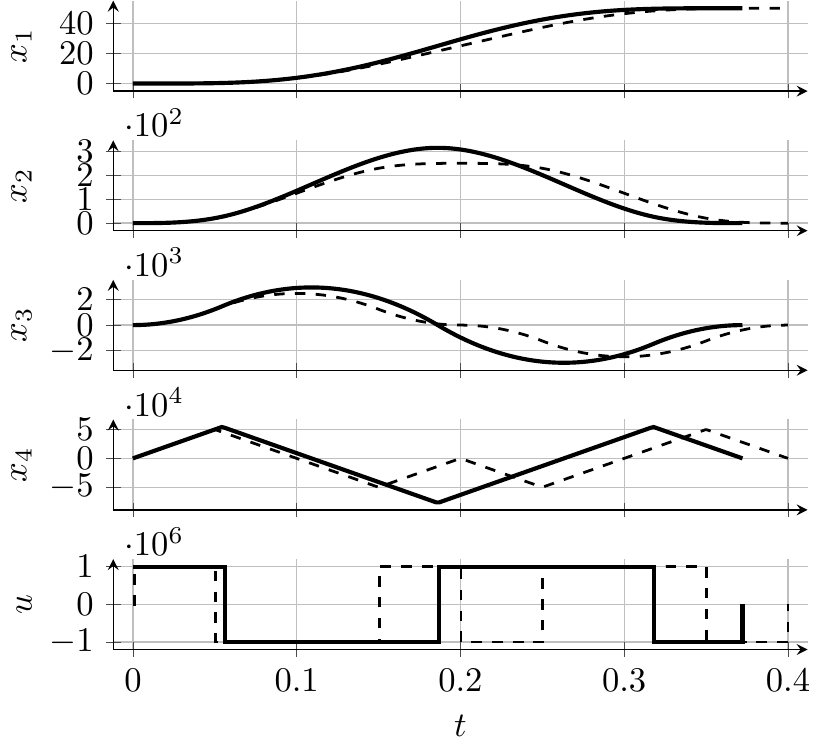}
	\vspace{-3mm}
	\caption{Time optimal (solid) fourth order trajectory and simplified one (dashed) with unnecessary saddle point in~$x_3$ (example from~\cite{Ezair2014}).}
	\label{C4saddlepoint}
	\vspace{-6mm}
\end{figure}

For at least up to~$n=3$, trajectories with non-rest initial and final conditions and asymmetric input and state bound constraints can be computed. Fig.~\ref{figc2trajectory} shows such a trajectory for~$n=3$. For~$n=4$ the case without state constraints, trajectories with non-rest initial and final conditions subject to asymmetric input bound constraints can as well be computed, see Fig.~\ref{figC3u}. For at least up to~$n=6$, trajectories with steady state initial and final conditions as well as symmetric input and state bound constraints can be planned. An example is illustrated in Fig.~\ref{figC5}. 
However, in order to compute such trajectories despite the many segments, all symmetry properties of the planning problem are exploited. Therefore, even if a state constraint is inactive, the corresponding contact condition (III.) is enforced which introduces unnecessary saddle points. This simplifies computations but time optimality is lost, as illustrated in Fig.~\ref{C4saddlepoint} for $n=4$. It shows the trajectory saddle with point in~$x_3$ in comparison with the time optimal solution without simplification. The loss of optimality has been pointed out in~\cite{Ezair2014}, as this simplification also applies for recursive planning approaches~\cite{Ezair2014,Knierim2012}.

\subsubsection{Computation times}
Typical computation times for Algorithm~\ref{algtraj} on a laptop with i7 processor @2.8~GHz for the general case with asymmetric input and state constrains and non-rest initial and final conditions are around 30~$\mu$s for~$n=2$ resp. around 700~$\mu$s for~$n=3$ and there is potential for improved computation times through further code optimization. Comparable values are achieved on a dSPACE MicroAutoBox II @900~MHz. In many applications, this opens up the possibility of online trajectory planning. For many applications, a planning in each time step is possible, e.g. for permanently re-planning based on joystick input. It may also be reasonable to run Algorithm~\ref{algtraj} in a slower task and evaluate an already computed trajectory precisely at arbitrary time values in real-time. Simplified rest-to-rest motions can be computed much faster and typically yield computation times around 15~$\mu$s for~$n=2$, 40~$\mu$s for~$n=3$, 80~$\mu$s for~$n=4$, 120~$\mu$s for~$n=4$ and 500~$\mu$s for~$n=6$.

\subsubsection{State space analysis}
\begin{figure}
	\centering
	\includegraphics[width=.95\linewidth]{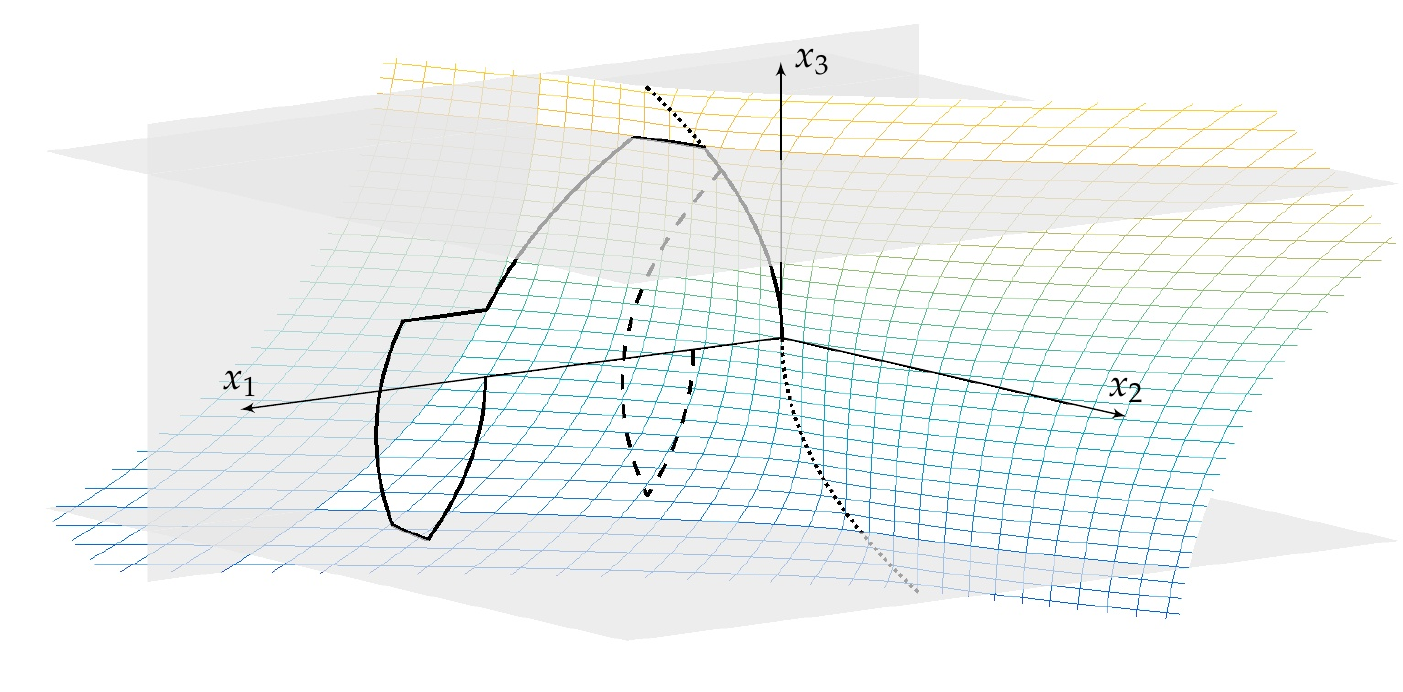}
	\vspace{-2mm}
	\caption{Two trajectories moving into the origin and switching surfaces for $n=3$. One trajectory (dashed) reaches no state constraints (type \textit{000}), the other one (solid) all constraints (type \textit{111}).}
	\label{fig:c23d}
	\vspace{-2mm}
\end{figure}

When analyzing the trajectory behavior for $n=3$ in state space, switching surfaces can be identified at which the trajectory transitions into the next segment. Fig.~\ref{fig:c23d} illustrates these surfaces together with two exemplary trajectories that move into the origin with profile type \textit{111} resp. \textit{000}. 
This property has been exploited in~\cite{Biagiotti2010} for trajectory generation by feedback control.

\subsubsection{Applications}
Exemplary use cases of this algebraic planning approach include the generation of valid reference states/outputs for feedback control and nominal control inputs for feedforward control, trajectory prediction for collision detection or brake distance prediction, three-dimensional motion or path planning for multi-axis systems, signal smoothing as a nonlinear filter, fast initial guess computation for trajectory optimization or realizing feedback control with finite time convergence by trajectory planning in flat coordinates with back-transformation. 

The proposed method has already been used in practical applications like reference generation of point-to-point motions for a high-precision measuring machine~\cite{Ringkowski2020GPobserver} and for the hydraulic clutch of an automotive gearbox in order to achieve fast filling using feedforward control based on integrator chain trajectories and differential flatness~\cite{Bauer2019}. The method is also used to plan three-dimensional load trajectories for sway-damping control of a full-size tower crane~\cite{Rauscher2020}.

\section{CONCLUSIONS}
This paper presents an efficient method to plan trajectories for integrator chains by algebraic computations without numerical approximation iterations or discretization. For at least up to system order three, a complete solution for arbitrary initial and final conditions as well as asymmetric input and state bound constraints is possible using this approach. For rest-to-rest trajectories or input bounds only, also sixth order trajectories can be planned. It generalizes planning procedures based on decision-trees and provides a methodical approach for algebraic planning based on and limited by a symbolic precomputation involving polynomial elimination. Exemplary trajectories are shown and implementation details are discussed in order to demonstrate the effectiveness of the proposed planning approach.


\begin{thebibliography}{99}
	\bibitem{Isidori2013}
	A.~Isidori, \emph{Nonlinear {Control} {Systems}}.\hskip 1em plus 0.5em minus
	0.4em\relax Springer Science, 2013.
	
	\bibitem{Fliess1990}
	M.~Fliess, ``Generalized controller canonical form for linear and nonlinear
	dynamics,'' \emph{{IEEE} Transactions on Automatic Control}, 1990.
	
	\bibitem{Biagiotti2008}
	L.~Biagiotti and C.~Melchiorri, \emph{{Trajectory
			{Planning} for {Automatic} {Machines} and {Robots}}}.\hskip 1em plus 0.5em
	minus 0.4em\relax Springer Science, 2008.
	
	\bibitem{Bazaz1999}
	S.~A. Bazaz and B.~Tondu, ``Minimum time on-line joint trajectory generator
	based on low order spline method for industrial manipulators,''
	\emph{Robotics and Autonomous Systems}, 1999.
	
	\bibitem{Macfarlane2003}
	S.~Macfarlane and E.~A. Croft, ``Jerk-bounded manipulator trajectory planning:
	design for real-time applications,'' \emph{IEEE Transactions on Robotics and
		Automation}, vol.~19, no.~1, pp. 42--52, Feb. 2003.
	
	\bibitem{Ezair2014}
	B.~Ezair, T.~Tassa, and Z.~Shiller, ``{Planning high
		order trajectories with general initial and final conditions and asymmetric
		bounds},'' \emph{{The International Journal of
			Robotics Research}}, vol.~33, 2014.
	
	\bibitem{Knierim2012}
	K.~L. Knierim and O.~Sawodny, ``Real-time trajectory generation for three-times
	continuous trajectories,'' in \emph{{ICIEA}}, 2012.
	
	\bibitem{Zanasi2002}
	R.~Zanasi and R.~Morselli, ``Third order trajectory generator satisfying
	velocity, acceleration and jerk constraints,'' in \emph{Proceedings of the
		{International} {Conference} on {Control} {Applications}}, 2002.
	
	\bibitem{Biagiotti2010}
	L.~Biagiotti and R.~Zanasi, ``Time-optimal regulation of a chain of integrators
	with saturated input and internal variables: an application to trajectory
	planning,'' \emph{IFAC Proceedings Volumes}, 2010.
	
	\bibitem{Liu2002}
	S.~Liu, ``An on-line reference-trajectory generator for smooth motion of
	impulse-controlled industrial manipulators,'' in \emph{7th {International}
		{Workshop} on {Advanced} {Motion} {Control}.}, 2002.
	
	\bibitem{Haschke2008}
	R.~Haschke, E.~Weitnauer, and H.~Ritter, ``{On-{Line}
		{Planning} of {Time}-{Optimal}, {Jerk}-{Limited} {Trajectories}},'' in
	\emph{{{IEEE}/{RSJ} {Int}. {Conf}. on {Intelligent}
			{Robots} and {Systems}}}, 2008.
	
	\bibitem{Besset2017}
	P.~Besset and R.~Béarée, ``{FIR} filter-based online jerk-constrained
	trajectory generation,'' \emph{Control Engineering Practice}, 2017.
	
	\bibitem{kroeger2010}
	T.~Kröger and F.~M. Wahl, ``Online trajectory generation: Basic concepts for
	instantaneous reactions to unforeseen events,'' \emph{IEEE Transactions on
		Robotics}, vol.~26, no.~1, pp. 94--111, Feb 2010.
	
	\bibitem{Lambrechts2005}
	P.~Lambrechts, M.~Boerlage, and M.~Steinbuch, ``Trajectory planning and
	feedforward design for electromechanical motion systems,'' \emph{Control
		Engineering Practice}, vol.~13, no.~2, pp. 145--157, Feb. 2005.
	
	\bibitem{Jeong2005}
	S.~Y. Jeong, Y.~J. Choi, P.~Park, and S.~G. Choi, ``Jerk limited velocity
	profile generation for high speed industrial robot trajectories,'' \emph{IFAC
		Proceedings Volumes}, vol.~38, no.~1, pp. 595--600, Jan. 2005.
	
	\bibitem{Buchberger1965}
	B.~Buchberger, ``{Ein Algorithmus zum Auffinden der Basiselemente des
		Restklassenringes nach einem nulldimensionalen Polynomideal},'' Ph.D.
	dissertation, {Leonold-Franzens-Universit\"at Innsbruck}, 1965.
	
	\bibitem{Walther2001}
	U.~Walther, T.~T. Georgiou, and A.~Tannenbaum, ``On the computation of
	switching surfaces in optimal control: {A} {Grobner} basis approach,''
	\emph{IEEE Transactions on automatic control}, 2001.
	
	\bibitem{Blaha2009}
	L.~Bláha, M.~Schlegel, and J.~Mošna, ``Optimal control of chain of
	integrators with constraints,'' in \emph{Proceedings of the 17th
		{International} {Conference} on {Process} {Control}}, vol.~9, 2009, pp.
	51--56.
	
	\bibitem{Blaha2014}
	L.~Bláha, ``Multi-axis {Time} {Synchronization} for {Uncoordinated} {Motion}
	{Planning} with {Hard} {Constraints},'' \emph{IFAC Proceedings Volumes},
	2014.
	
	\bibitem{Ringkowski2020GPobserver}
	M.~Ringkowski, O.~Sawodny, S.~Hartlieb, T.~Haist, and W.~Osten, ``Estimating
	dynamic positioning errors of coordinate measuring machines,''
	\emph{Mechatronics}, vol.~68, p. 102383, jun 2020.
	
	\bibitem{Bauer2019}
	M.~Bauer and O.~Sawodny, ``Near time-optimal two-staged flatness based
	feed-forward control of a clutch filling process,''
	\emph{{IFAC}-{PapersOnLine}}, vol.~52, no.~15, pp. 205--210, 2019.
	
	\bibitem{Rauscher2018}
	F.~Rauscher, S.~Nann, and O.~Sawodny, ``Motion control of an overhead crane
	using a wireless hook mounted {IMU},'' in \emph{{ACC}}, 2018.
	
	\bibitem{Rauscher2020}
	F.~Rauscher and O.~Sawodny, ``Modeling and control of tower cranes with elastic
	structure,'' \emph{{IEEE} Transactions on Control Systems Technology}, pp.
	1--16, 2020.
	
	\bibitem{feld1965optimal}
	A.~Feld’baum, ``Optimal control systems,'' \emph{Mathematics in science and
		engineering. New York: Academic Press}, 1965.
	
	\bibitem{Wu1989}
	W.~Wentsun, ``A {Zero} {Structure} {Theorem} for {Polynomial}-equations-solving
	and {Its} {Applications},'' in \emph{Proceedings of the {European}
		{Conference} on {Computer} {Algebra}}.\hskip 1em plus 0.5em minus 0.4em\relax
	Springer-Verlag, 1989.
	
	\bibitem{kalkbrener1993generalized}
	M.~Kalkbrener, ``A generalized euclidean algorithm for computing triangular
	representations of algebraic varieties,'' \emph{Journal of Symbolic
		Computation}, vol.~15, no.~2, pp. 143--167, 1993.
	
	\bibitem{moller1984upper}
	H.~M. M{\"o}ller and F.~Mora, ``Upper and lower bounds for the degree of
	gr{\"o}bner bases,'' in \emph{International Symposium on Symbolic and
		Algebraic Manipulation}.\hskip 1em plus 0.5em minus 0.4em\relax Springer,
	1984, pp. 172--183.
	
	\bibitem{wilkinson1963}
	J.~Wilkinson, \emph{Rounding Errors in Algebraic Processes}.\hskip 1em plus
	0.5em minus 0.4em\relax Prentice-Hall, Englewood Cliffs, 1963.
\end{thebibliography}
\end{document}